%% file: acl_latex.tex
\documentclass[11pt]{article}

\usepackage[preprint]{acl}

\usepackage{times}
\usepackage{latexsym}

\usepackage[T1]{fontenc}

\usepackage[utf8]{inputenc}

\usepackage{microtype}

\usepackage{inconsolata}

\usepackage{graphicx}

\usepackage{amssymb}
\usepackage{amsmath}
\usepackage{amssymb}
\usepackage{mathtools}
\usepackage{amsthm}
\usepackage{multirow}
\usepackage{booktabs} 
\usepackage{multirow} 
\usepackage{graphicx} 
\usepackage{makecell} 
\usepackage{graphicx}
\usepackage{subcaption}
\usepackage{xcolor}
\usepackage{listings}
\usepackage{tcolorbox}
\usepackage{tabularx}
\usepackage{threeparttable}
\usepackage[table]{xcolor} 

\tcbuselibrary{listings, skins, breakable}

\newtcblisting{promptbox}[2][]{
    colback=gray!5,       
    colframe=gray!60!black, 
    arc=2pt, outer arc=2pt, 
    boxrule=0.8pt,        
    listing only,         
    listing options={
        basicstyle=\ttfamily\small, 
        breaklines=true,      
        breakatwhitespace=true, 
        columns=fullflexible, 
        keepspaces=true,      
        language=,            
        aboveskip=0pt,
        belowskip=0pt,
        showstringspaces=false
    },
    title={\textbf{#2}},  
    #1
}

\title{Beyond Transcription: Unified Audio Schema for Perception-Aware AudioLLMs}

\author{
    \textbf{Linhao Zhang\textsuperscript{1}\thanks{Equal contribution.}}\textmd{,}
    \textbf{Yuhan Song\textsuperscript{2}\footnotemark[1]}\textmd{,}
    \textbf{Aiwei Liu\textsuperscript{1}}\textmd{,}
    \textbf{Chuhan Wu\textsuperscript{1}}\textmd{,}
    \\ 
    \textbf{Sijun Zhang\textsuperscript{1}},
    \textbf{Wei Jia\textsuperscript{1}},
    \textbf{Yuan Liu\textsuperscript{1}},
    \textbf{Houfeng Wang\textsuperscript{2}},
    \textbf{Xiao Zhou\textsuperscript{1}}
    \\
    \\
    \textsuperscript{1}Basic Model Technology Center, WeChat AI, Tencent Inc.
    \\
    \textsuperscript{2}State Key Laboratory of Multimedia Information Processing, \\School of Computer Science, Peking University
}

\begin{document}
\maketitle
\begin{abstract}

Recent Audio Large Language Models (AudioLLMs) exhibit a striking performance inversion: while excelling at complex reasoning tasks, they consistently underperform on fine-grained acoustic perception. We attribute this gap to a fundamental limitation of ASR-centric training, which provides precise linguistic targets but implicitly teaches models to suppress paralinguistic cues and acoustic events as noise. To address this, we propose Unified Audio Schema (UAS), a holistic and structured supervision framework that organizes audio information into three explicit components—Transcription, Paralinguistics, and Non-linguistic Events—within a unified JSON format. This design achieves comprehensive acoustic coverage without sacrificing the tight audio-text alignment that enables reasoning. We validate the effectiveness of this supervision strategy by applying it to both discrete and continuous AudioLLM architectures. Extensive experiments on MMSU, MMAR, and MMAU demonstrate that UAS-Audio yields consistent improvements, boosting fine-grained perception by 10.9\% on MMSU over the same-size state-of-the-art models while preserving robust reasoning capabilities. Our code and model are publicly available at \url{https://github.com/Tencent/Unified_Audio_Schema}.


\end{abstract}

\section{Introduction}
\input{introduction}

\section{Unified Audio Schema (UAS)}
\input{uas}

\section{UAS-Audio}

\input{uas_audio}

\section{Experimental Setup}
\input{setup}

\section{Results}
\input{results}

\section{Related Work}
\input{related_work}

\section{Conclusion}
\input{conclusion}

\section*{Limitations}

Despite the promising results of UAS-Audio in bridging the perception-reasoning gap, we acknowledge certain limitations in our current study that point to directions for future research:

\begin{itemize}

\item \textbf{Linguistic Diversity:} Our current experimental validation primarily focuses on high-resource languages (English and Chinese) due to the composition of mainstream benchmarks like MMSU and Seed-TTS. While the UAS framework is language-agnostic by design, its efficacy on low-resource or code-switching scenarios remains to be verified in future work.

 \item \textbf{Complex Overlapping Speech:} While UAS effectively handles background environmental events, our current schema focuses on the primary speaker for paralinguistic analysis. Scenarios involving highly overlapping speech (e.g., the cocktail party problem) with multiple active speakers requiring simultaneous paralinguistic disentanglement are not fully covered in this iteration and warrant further investigation.

 \end{itemize}









\bibliography{custom}

\appendix
\input{appendix}


\end{document}

%% file: introduction.tex
\begin{figure*}[t]
    \centering
    \begin{subfigure}[b]{0.49\textwidth}
        \centering
        \includegraphics[width=\linewidth]{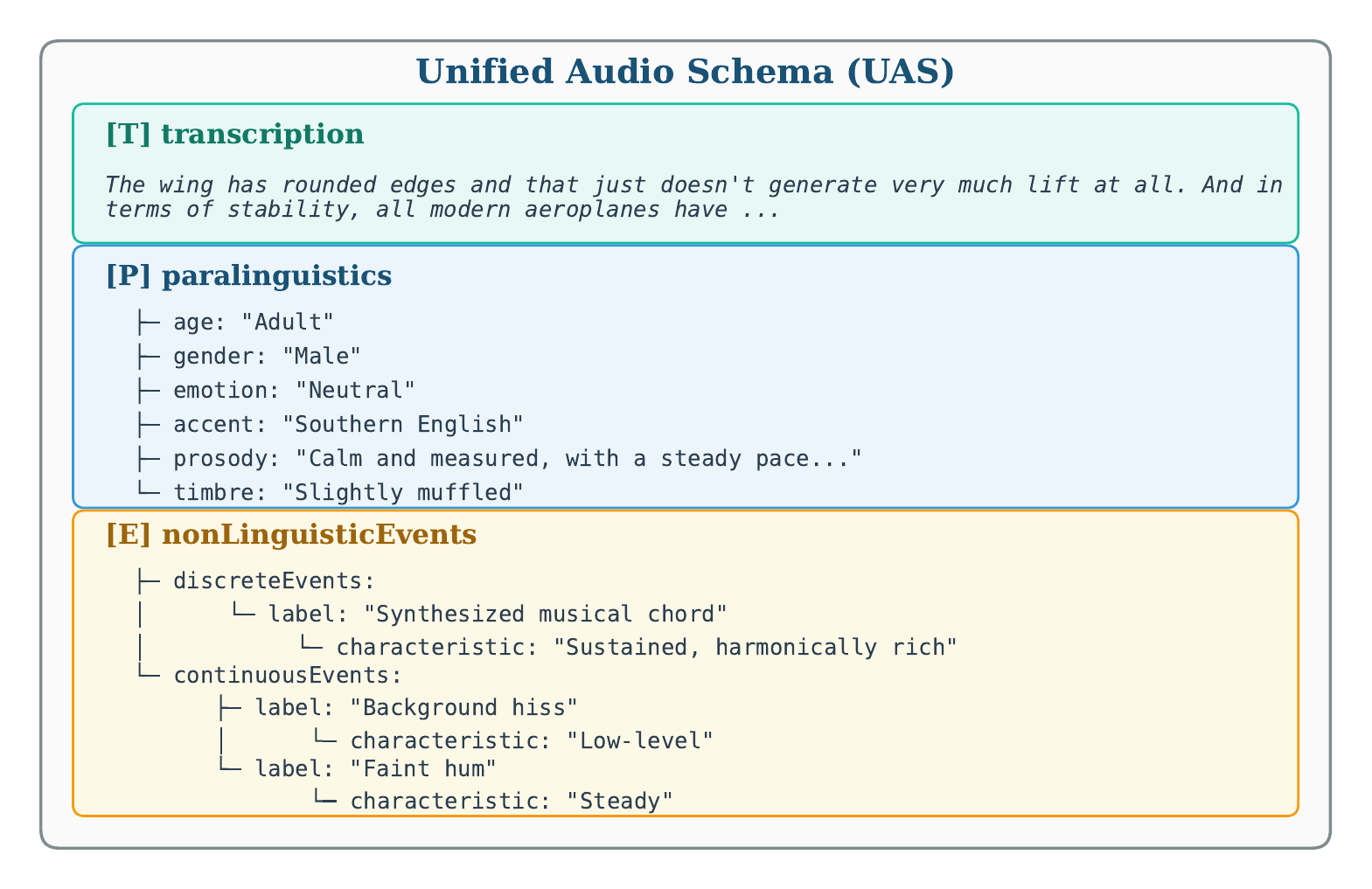}
        \caption{Unified Audio Schema (UAS)}
        \label{fig:uas_schema}
    \end{subfigure}
    \hfill
    \begin{subfigure}[b]{0.49\textwidth}
        \centering
        \includegraphics[width=\linewidth]{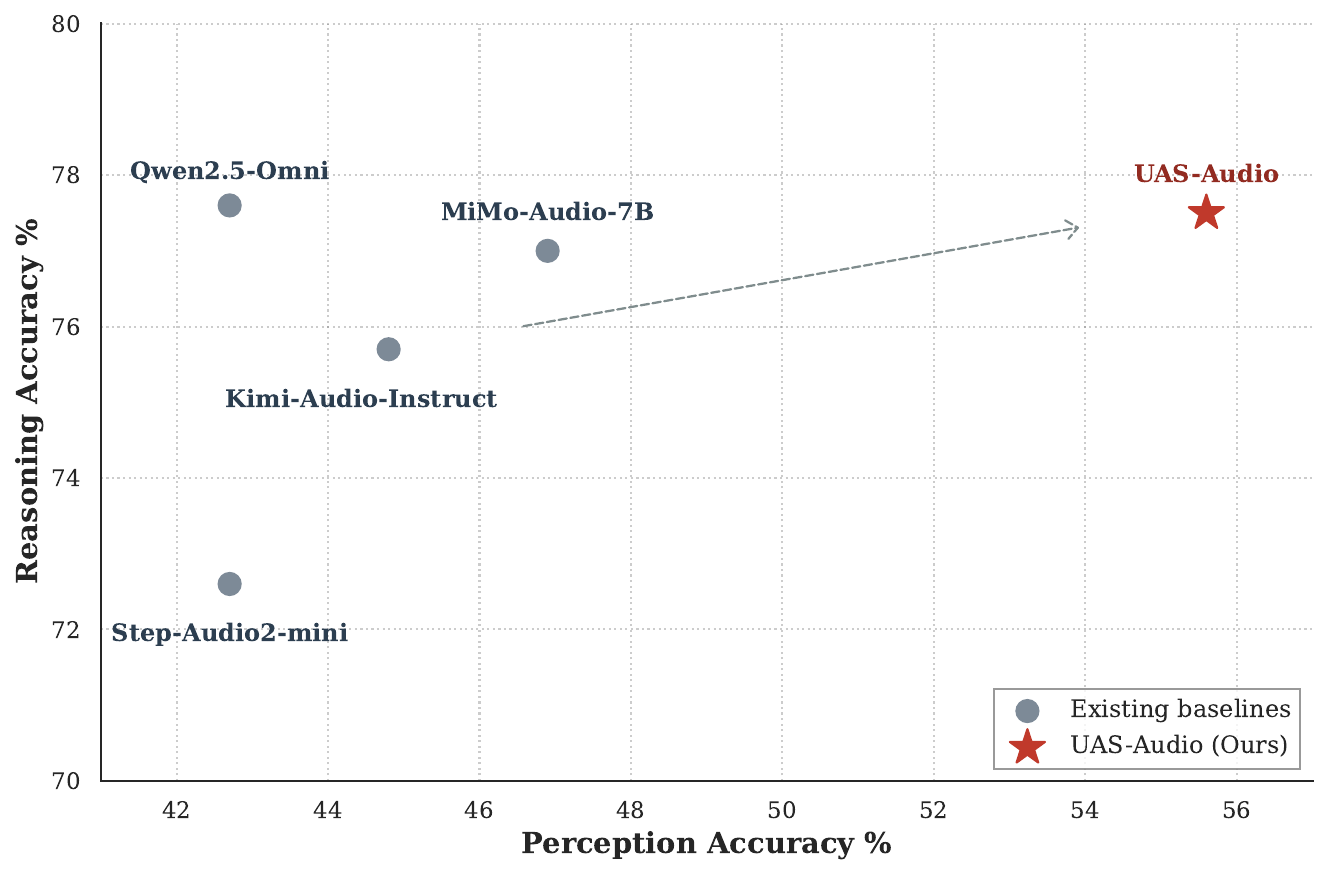}
        \caption{Performance of UAS-Audio}
        \label{fig:uas_performance}
    \end{subfigure}
    \caption{\textbf{Overview of the Unified Audio Schema (UAS) and evaluation results.} 
    (a) UAS structures audio information into three components: Transcription, Paralinguistics, and Non-linguistic Events. 
    (b) Reasoning vs. Perception accuracy on MMSU. UAS-Audio significantly enhances perception while maintaining robust reasoning.}
    \label{fig:overall_architecture}
\end{figure*}

Recent Audio Large Language Models (AudioLLMs) present a striking paradox: models capable of complex reasoning often fail at elementary auditory perception. While they excel on reasoning-heavy benchmarks, they struggle to reliably identify speaker traits, emotion, prosody, or even simple non-linguistic acoustic events~\cite{yang2024airbenchbenchmarkinglargeaudiolanguage,wang2025mmsu, kwon2025m3sluevaluatingspeakerattributedreasoning,zhang2025wildspeech}. For instance, on the MMSU benchmark, current AudioLLMs achieve approximately 70\% accuracy on complex reasoning tasks yet drop sharply to around 40\% on fundamental perception tasks. Such perceptual blind spots lead to practical failures: a model may correctly transcribe "I'm fine" while completely missing the trembling voice that signals distress, or fail to notice a door slam that signals an abrupt end to the interaction.

Despite the rapid scaling of both language backbones and audio encoders, these perceptual failures persist across different model sizes and architectures. This persistence suggests that the underlying issue is unlikely to be solely a result of insufficient model capacity or architectural limitations. Instead, it points to a more systemic factor shared by most existing AudioLLMs: how audio information is supervised during training.

Most existing models rely heavily on automatic speech recognition (ASR) as their training signal, using text transcription as the primary interface between audio and language. While effective for semantic alignment, ASR is inherently selective: to recover canonical text, it deliberately normalizes away prosody, speaker identity, emotion, and acoustic context.
This training objective creates a fundamental asymmetry---models are consistently rewarded for reasoning about \textit{what} is said, while being implicitly discouraged from attending to \textit{how} it is said or what else occurs acoustically.
As a result, perception is not merely under-trained---it is systematically de-emphasized.

Motivated by this perspective, we introduce the \textbf{Unified Audio Schema (UAS)}, a structured textual representation that decomposes audio into three complementary components (Figure~\ref{fig:uas_schema}): \textit{Transcription} captures the spoken content; \textit{Paralinguistics} encodes speaker-level attributes such as emotion, age, gender, accent, prosody, and timbre; and \textit{Non-linguistic Events} describes the acoustic context, including both discrete sounds (e.g., door slams, laughter) and continuous background conditions (e.g., ambient noise, music). This decomposition follows the classical taxonomy of speech information~\citep{laver1994principles} and is designed to expose perceptual dimensions explicitly rather than implicitly.
Leveraging this structured format, UAS enables AudioLLMs to retain perceptual information without sacrificing semantic alignment.

Crucially, this schema does not require expensive manual annotation. We demonstrate that UAS training data can be automatically synthesized at scale from existing corpora using off-the-shelf models, converting standard ASR datasets into rich, perception-aware supervision.

Leveraging this scalable pipeline, we validate the effectiveness of UAS on both continuous ~\cite{Qwen2.5-Omni, wu2025stepaudio2technicalreport} and discrete ~\cite{zeng2024glm} architectures. Experiments confirm that restructuring supervision yields consistent gains across both paradigms. As shown in Figure~\ref{fig:uas_performance}, UAS-Audio achieves an 11\% absolute improvement in perception accuracy over state-of-the-art baselines on MMSU, while strictly preserving reasoning capabilities. Beyond perception, UAS-Audio demonstrates robust generalization across diverse audio domains, achieving state-of-the-art performance on the MMAR reasoning benchmark (60.1\%) and competitive results on MMAU, securing the highest average score (65.2\%) across all three benchmarks.

In summary, our contributions are threefold:
\begin{itemize}
    \item We identify that the perceptual weakness of current AudioLLMs stems from ASR-centric supervision, which systematically de-emphasizes paralinguistic and non-linguistic information during training.
    
    \item We propose the Unified Audio Schema (UAS), a structured representation that explicitly decomposes audio into transcription, paralinguistics, and non-linguistic events, along with a scalable pipeline to synthesize UAS annotations using off-the-shelf expert models.
    
    \item We train UAS-Audio and demonstrate consistent improvements across both continuous and discrete AudioLLM architectures, achieving an absolute 11\% gain in perception accuracy on MMSU benchmark while preserving reasoning performance.
\end{itemize}

%% file: uas.tex
\label{sec:uas}

\begin{figure*}[t]
    \centering
    \includegraphics[width=\textwidth]{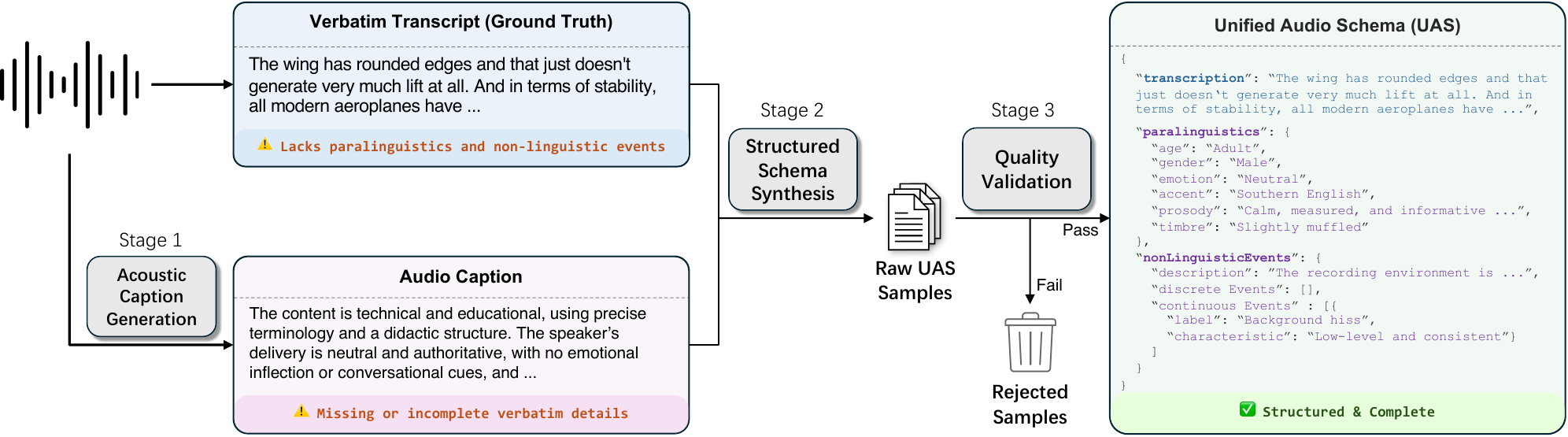}
    \caption{Overview of the UAS Data Generation Pipeline. (Stage 1) Acoustic captions are generated to capture paralinguistic and environmental context. (Stage 2) A structured synthesizer merges these captions with ground-truth transcripts.  (Stage 3) An automated validation stage filters hallucinations and enforces logical consistency, yielding a final UAS that is `Structured \& Complete'}
    \label{fig:overview}
\end{figure*}

\subsection{Schema Definition and Rationale}

Our schema design is grounded in the semiotic framework of speech signals defined by ~\citet{laver1994principles}. According to Laver, speech is not a monolithic stream but a composite of three information layers: (1) the Linguistic Layer, conveying the abstract semantic message; (2) the Paralinguistic Layer, encompassing voluntary vocal variations (e.g., tone, prosody) that signal attitude; and (3) the Extralinguistic Layer, containing indexical features identifying biological constraints (e.g., age, gender) and the physical environment.

Building upon this hierarchy, we aim to preserve the full spectrum of acoustic nuance by explicitly disentangling the signal's semantic and non-semantic components. We map Laver’s theoretical layers into a practical annotation schema organized along three functional dimensions:

\paragraph{Transcription.}
Corresponding to the linguistic layer, this field preserves the verbatim speech content with 100\% linguistic fidelity equivalent to ASR output. This field ensures that UAS never sacrifices the semantic precision compared to standard ASR.
    
\paragraph{Paralinguistics.}
Capturing the paralinguistic layer and the speaker-intrinsic aspects of the extralinguistic layer, this field consists of structured attributes describing \textit{how} speech is produced. It is organized into six explicit subfields: (1) \textit{Age}: Speaker age group (e.g., ``Adult'', ``Child'', ``Elderly''); (2) \textit{Gender}: Speaker gender (e.g., ``Male'', ``Female''); (3) \textit{Emotion}: Emotional state (e.g., ``Neutral'', ``Happy'', ``Angry'', ``Sad''); (4) \textit{Accent}: Regional or linguistic accent (e.g., ``Southern American English''); (5) \textit{Prosody}: Speaking style description including pace, intonation, and rhythm; and (6) \textit{Timbre}: Voice quality characteristics describing the acoustic texture of the speaker's voice (e.g., ``Slightly muffled'', ``Clear and resonant'').
    
\paragraph{Non-linguistic Events.}
Representing the environmental aspect of the extralinguistic layer, this field gives acoustic information beyond speech, capturing the auditory scene through three subfields: (1) \textit{Description}: Overall characterization of the recording environment; (2) \textit{Discrete Events}: Identifiable sound occurrences with clear temporal boundaries (e.g., door slam); and (3) \textit{Continuous Events}: Ambient sounds persisting throughout the audio (e.g., engine rumble). Each discrete or continuous event is annotated with a label and characteristic (e.g., ``stylus click'' with ``sharp, high-pitched'', ``Background hiss'' with ``low-level and consistent'').

It is important to note that UAS supports both speech and non-speech data (e.g., pure environment sounds or music). For audio segments containing no human speech, the ``transcription'' field and all relevant paralinguistic subfields are set to ``null''.

This structured decomposition offers critical advantages beyond mere organization. 
First, it facilitates \textbf{disentangled learning} by transforming the implicit task of ``holistic understanding'' into explicit subtasks. This prevents feature conflation, forcing the model to distinguish \textit{what} is said from \textit{how} it is said. 
Second, the schema introduces \textbf{syntactic invariance}. Unlike unstructured captions that suffer from high entropic variability (i.e., many ways to describe the same sound), UAS provides a consistent, low-entropy supervision target, thereby reducing learning difficulty and stabilizing optimization. 
Finally, it ensures \textbf{programmatic accessibility}. The rigorous JSON format bridges the gap between the probabilistic nature of LLMs and the deterministic requirements of software interfaces, allowing downstream applications to reliably utilize acoustic attributes without complex parsing.

\subsection{Scalable UAS Data Generation Pipeline}
\label{sec:pipeline}

To operationalize UAS at scale, we develop an automated pipeline that transforms existing ASR corpora into UAS-formatted supervision. The pipeline proceeds in three stages, as illustrated in Figure~\ref{fig:overview}.

\paragraph{Stage 1: Acoustic Caption Generation.}
Given raw audio samples with their original transcriptions, we first employ a caption model ~\cite{Qwen3-Omni} as an acoustic captioner to generate rich descriptions of paralinguistic attributes and environmental sounds. The captioner is prompted to describe speaker characteristics (age, gender, emotion, accent, prosody, timbre), and non-speech acoustic events (e.g., background environment, discrete sounds, continuous ambient sounds or music) in detail. This stage extracts the acoustic information that ASR supervision inherently discards.

\paragraph{Stage 2: Structured Schema Synthesis.}
The generated acoustic captions are then combined with ground-truth transcriptions and processed through an LLM to synthesize structured UAS annotations. We design a carefully hand-crafted prompt (detailed in Appendix~\ref{app:prompts}) that instructs the LLM to: (1) preserve the original transcription verbatim in the designated field; (2) extract and normalize paralinguistic attributes from the caption into the predefined categories (age, gender, emotion, accent, prosody, timbre); and (3) organize non-linguistic event descriptions into the hierarchical schema with discrete and continuous event separation. This synthesis step ensures both semantic fidelity (via ground-truth transcription) and acoustic completeness (via caption-derived attributes).

\paragraph{Stage 3: Quality Validation.} To ensure strict data reliability, we implement a multi-level automated validation pipeline: (1) Ontology Constraints Enforcement: All categorical fields (e.g., emotion, gender) are verified against a predefined closed-set vocabulary, discarding synonyms or open-ended hallucinations from the LLM. (2) Transcription Integrity: Exact string matching with the ground truth ensures zero semantic loss. (3) Logical Consistency Filtering: Rule-based checks resolve inter-field conflicts, such as strictly mapping empty transcriptions to “null” paralinguistic fields, and rejecting samples where gender attributes contradict acoustic timbre (e.g., “Male” label with “High-pitched feminine” description). (4) Duration-Content Alignment: Heuristic filters discard samples where the description’s complexity disproportionately exceeds the audio duration, reducing the risk of generative hallucination.

\paragraph{Human Verification of Data Quality.}
To verify the automated pipeline, we manually audited $N=400$ random samples across speech, music, and environmental sounds. The results demonstrate high reliability, with most paralinguistic and environmental attributes achieving a mean accuracy of over 95\%. Detailed methodology and per-field accuracy scores are provided in Appendix~\ref{app:human_eval}.

\subsection{UAS-QA}
Beyond raw UAS annotations, we additionally synthesize a \textbf{UAS-QA} dataset to train the model to leverage structured acoustic knowledge for downstream tasks. Based on UAS annotations, we automatically generate three types of question-answer pairs: (1) \textbf{Direct QA}: Questions querying specific UAS fields (e.g., ``What is the speaker's emotion?'' $\rightarrow$ ``Neutral''; ``What is the speaker's accent?'' $\rightarrow$ ``Southern American English''); (2) \textbf{Multiple Choice}: Questions with candidate options derived from the UAS attribute vocabulary (e.g., ``What is the speaker's age group? A. Child B. Adult C. Elderly'' $\rightarrow$ ``B. Adult''); and (3) \textbf{Yes/No Questions}: Binary verification questions (e.g., ``Is the speaker male?'' $\rightarrow$ ``Yes''; ``Are there discrete sound events in this audio?'' $\rightarrow$ ``No'').

This diverse question format ensures comprehensive coverage of all UAS fields---transcription, paralinguistics, and non-linguistic events---while providing explicit supervision signals that encourage the model to attend to fine-grained acoustic details during inference. 

%% file: uas_audio.tex
\subsection{Overview}
\label{sec:model}

\begin{figure}[t]
  \centering
  \includegraphics[width=0.48\textwidth]{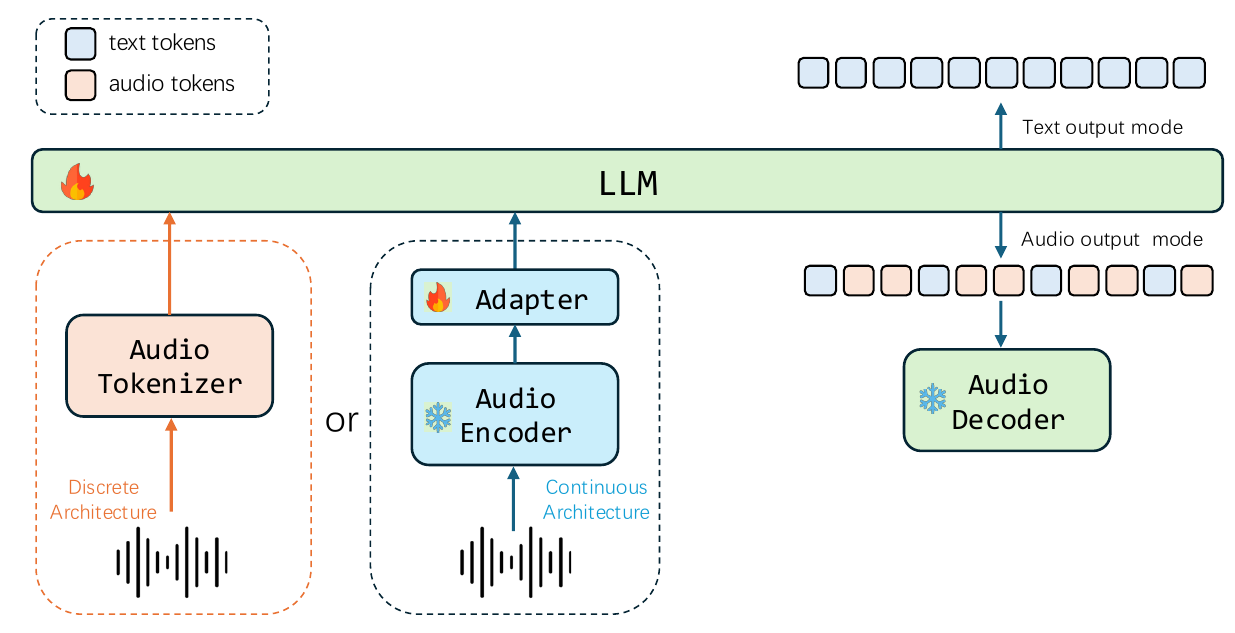}
  \caption{\textbf{Overview of UAS-Audio architectures}. Audio input is processed via \textit{either} discrete tokenization \textit{or} continuous encoding (left). 
The LLM generates text-only or bi-modal outputs (right). }
  \label{fig:uas_audio_arc}
\end{figure}

To validate the effectiveness of UAS supervision, we develop \textbf{UAS-Audio} (Figure ~\ref{fig:uas_audio_arc}), an AudioLLM designed to address the perception–reasoning imbalance identified
in existing models.
In this section, we focus on the continuous architecture of UAS-Audio, whose
design aims to preserve the strong reasoning capabilities of current AudioLLMs
while substantially enhancing fine-grained acoustic perception---achieving holistic
audio understanding without the trade-offs inherent in ASR-centric or caption-based
training paradigms. The discrete variant (denoted as UAS-Audio-D), which follows a similar design philosophy with architecture-specific adaptations, is detailed in
Appendix~\ref{app:discrete_training}.

Following the successful paradigm established by recent
AudioLLMs~\cite{Qwen2.5-Omni,wu2025stepaudio2technicalreport}, the continuous
UAS-Audio model adopts a four-component framework that has proven effective for
audio–language modeling:
(1) an \textbf{Audio Encoder} that transforms raw waveforms into continuous
representations;
(2) a \textbf{Projection Layer} that aligns audio representations with the language
model’s embedding space;
(3) a \textbf{Large Language Model} backbone that performs reasoning over combined
audio–text inputs; and
(4) a \textbf{Speech Decoder} based on the flow matching architecture~\cite{fm2023},
which converts audio tokens into mel-spectrograms that are subsequently transformed
into waveforms using a HiFi-GAN vocoder~\cite{kong2020hifi}.

We next describe the training pipeline for this continuous architecture.

\subsection{Training Pipeline}
We follow the standard multi-stage alignment protocol established in \citet{wu2025stepaudio2technicalreport,Qwen2.5-Omni}. We do not introduce new modules or specialized loss functions; we simply plug in the UAS data.

\paragraph{Stage 1: Discrete Token Alignment.}

Although our primary architecture leverages a continuous audio encoder (plus adapter) for high-fidelity input processing, discrete audio tokens remain essential for enabling the LLM to perform speech generation. To establish this output capability, we extend the LLM's vocabulary with discrete acoustic codes derived from StableToken~\cite{song2026stabletoken}. This stage aligns text and audio representations via Automatic Speech Recognition (ASR) and Text-to-Speech (TTS) tasks, effectively equipping the model with the interface to autoregressively predict audio segments for the speech decoder. During this process, only the embedding layer and the LLM head are trainable, while all other model parameters remain frozen.

\paragraph{Stage 2: Audio-LLM Adaptation.}
The second stage adapts the audio encoder to the pretrained LLM for cross-modal alignment. We train exclusively on UAS annotation data, with the LLM backbone and audio encoder frozen and only the projection layer updated. By introducing structured acoustic understanding at the outset, this stage prevents the model from developing ASR-centric representations that would later need to be ``unlearned'' to accommodate paralinguistic information.

\paragraph{Stage 3: Full Instruction Tuning.}
The third stage performs comprehensive instruction tuning across diverse audio understanding and generation tasks. We unfreeze all model parameters except the audio encoder and train on a mixture of: (1) \textbf{Foundational audio data}, following prior work~\cite{Qwen2.5-Omni,wu2025stepaudio2technicalreport}, including ASR and TTS tasks; (2) \textbf{UAS annotation} for generating complete UAS JSON from audio input; and (3) \textbf{UAS-QA} for answering questions about specific acoustic attributes across all UAS fields.

This diverse task mixture ensures that the model develops comprehensive audio intelligence spanning perception, reasoning, and generation. The inclusion of both UAS annotation and UAS-QA data---as validated by our ablation study---provides complementary supervision: UAS annotation teaches \textit{what} to perceive, while UAS-QA teaches \textit{how} to apply this knowledge.

\paragraph{Stage 4: GRPO.}
Following previous work ~\cite{wu2025stepaudio2technicalreport}, we utilize Group Relative Policy Optimization (GRPO)~\cite{shao2024deepseekmathpushinglimitsmathematical,li2025reinforcement} to further enhance the model's capabilities.

%% file: setup.tex
\label{sec:setup}

In this section, we detail the experimental setup, including the evaluation benchmarks, baseline models, and implementation details.

\subsection{Evaluation Benchmarks}

To comprehensively evaluate audio understanding and reasoning capabilities, we utilize benchmarks that collectively assess perception, reasoning, and generation abilities.

\paragraph{Audio Understanding.} We evaluate audio understanding using three benchmarks. \textbf{MMSU}~\cite{wang2025mmsu} is a comprehensive benchmark for understanding and reasoning in spoken language, comprising 5,000 audio-question-answer triplets across 47 tasks. Notably, MMSU is divided into \textit{perception} and \textit{reasoning} subsets, making it particularly relevant to our work as it directly measures the paralinguistic perception abilities that UAS targets. \textbf{MMAU}~\cite{sakshi2024mmaumassivemultitaskaudio} evaluates multimodal audio understanding on tasks requiring expert-level knowledge and complex reasoning, comprising 10k audio clips with human-annotated QA pairs spanning speech, environmental sounds, and music. \textbf{MMAR}~\cite{ma2025mmar} evaluates deep reasoning capabilities of Audio-Language Models across multi-disciplinary tasks, containing 1,000 curated audio-question-answer triplets where each item demands multi-step reasoning beyond surface-level understanding. 

\paragraph{Audio Generation.} We evaluate audio generation capabilities using \textbf{Seed-TTS}~\cite{anastassiou2024seed} on both Chinese (Seed-zh) and English (Seed-en) test sets. This benchmark assesses the model's fundamental text-to-speech synthesis ability, ensuring that UAS training does not compromise basic generation quality.

\subsection{Baselines}

We compare against three state-of-the-art Audio-Language Models with 7B-scale language model backbones: Qwen2.5-Omni~\cite{Qwen2.5-Omni}, a unified multimodal model capable of perceiving and generating across text, images, audio, and video; Kimi-Audio~\cite{ding2025kimi}, an audio-language model with strong audio understanding and generation capabilities; and Step-Audio2-mini~\cite{wu2025stepaudio2technicalreport}, a compact yet powerful model designed for efficient audio understanding.

Results are cited from \citet{coreteam2025mimoaudio} and we adopt the same evaluation framework.

\subsection{Implementation Details}

We build UAS-Audio upon the Qwen2.5-7B~\cite{Qwen2.5-Omni} language model backbone with AuT (Audio Transformer)~\cite{Qwen3-Omni} as our audio encoder, using a linear projection layer to align audio representations with the language model's embedding space. We use the AdamW optimizer with cosine learning rate scheduling and linear warmup across all stages.  A comprehensive list of the specific datasets used for training is detailed in Appendix~\ref{app:datasets}. Detailed hyperparameter configurations for each training stage are provided in Appendix~\ref{app:hyperparameters}.

While UAS serves as a dense supervision signal during training, we do not mandate the generation of full UAS JSONs during inference. Instead, we adopt a \textbf{task-specific prompting strategy} to align with standard evaluation protocols: we use standard transcription prompts for ASR, synthesis prompts for TTS, and discriminative prompts (e.g., multiple-choice) for understanding benchmarks like MMSU. This approach ensures our method remains compatible with existing metrics and incurs no additional latency overhead compared to task-specific baselines.

%% file: results.tex
\subsection{Main Results}
\begin{table*}[t]
\centering

\setlength{\tabcolsep}{2.5pt}
\renewcommand{\arraystretch}{1.15}

\resizebox{\textwidth}{!}{%
\begin{tabular}{l ccc cccc cccc c}
\toprule

\multirow{2}{*}{\textbf{Model}} 
& \multicolumn{3}{c}{\textbf{MMSU}} 
& \multicolumn{4}{c}{\textbf{MMAR}} 
& \multicolumn{4}{c}{\textbf{MMAU}} 
& \multirow{2}{*}{\textbf{Avg.}} \\

\cmidrule(lr){2-4} \cmidrule(lr){5-8} \cmidrule(lr){9-12}

& Perception & Reasoning & Overall 
& Speech & Sound & Music & Overall
& Speech & Sound & Music & Overall
& \\

\midrule
\multicolumn{13}{c}{\textit{\textbf{Discrete Input Architecture}}} \\
GLM-4-Voice
& 11.04 & 16.16 & 13.30

& 34.35 & 29.70 & 19.90 & 29.60
& 35.44 & 27.63 & 27.84 & 30.30
& 24.4 \\

\textbf{UAS-Audio-D}
& 31.32 & 48.55 & 39.66
& 44.56 & 40.00 & 36.89 & 43.30
& 46.25 & 59.16 & 62.57 & 56.00
& 44.2 \\

\midrule
\multicolumn{13}{c}{\textit{\textbf{Continuous Input Architecture}}} \\
Kimi-Audio 
& \underline{44.8} & 75.7 & 59.8            
& 58.5 & 49.7 & 33.0 & 48.0      
& 62.2 & 75.7 & 66.8 & 68.2      
& 58.7 \\                              

Qwen2.5-Omni
& 42.7 & \textbf{77.6} & \underline{58.1}            
& \underline{59.9} & \underline{58.8} & \underline{40.8} & \underline{56.7}      
& \textbf{70.6} & \underline{78.1} & 65.9 & \underline{71.5}      
& \underline{62.1} \\                              

Step-Audio2
& 42.9 & 73.2 & 57.6            
& 61.2 & 54.6 & 42.2 & 56.8      
& \underline{68.2} & \textbf{79.3} & \underline{68.4} & \textbf{72.7}      
& 61.9 \\                              

\textbf{UAS-Audio} 
& \textbf{55.7}~\textcolor{blue}{(+10.9)} & \underline{77.4} & \textbf{66.2}          
& \textbf{66.0} & \textbf{58.8} & \textbf{45.2} & \textbf{60.1}      
& 67.0 & 70.0 & \textbf{71.3} & 69.4      
& \textbf{65.2} \\    

\bottomrule
\end{tabular}%
}
\caption{Unified evaluation results on MMSU, MMAR, and MMAU benchmarks. The models are categorized by their audio input representation (Discrete vs. Continuous). \textbf{Bold} denotes the best result; \underline{underline} denotes the second best. UAS-Audio-D denotes the discrete-input variant of UAS-Audio.}
\label{tab:final_unified_results}
\end{table*}

Table~\ref{tab:final_unified_results} presents comprehensive evaluation results across three benchmarks. Our UAS-Audio achieves the highest overall average of 65.2\%, outperforming all baselines by a substantial margin.

\paragraph{Perception-Reasoning Trade-off.}
The most striking finding emerges from the MMSU benchmark, which explicitly decouples perception and reasoning capabilities. UAS-Audio achieves an accuracy of 55.7\% on perception tasks, a remarkable \textbf{+10.9\%} absolute improvement over the best baseline (Kimi-Audio at 44.8\%). Crucially, this perception gain does not come at the cost of reasoning: UAS-Audio maintains competitive reasoning accuracy (77.4\%), only 0.2\% below the top-performing Qwen2.5-Omni (77.6\%). This result directly validates our hypothesis that ASR-centric training creates a perception bottleneck that UAS effectively addresses, while the structured schema format preserves the model's reasoning capabilities.

\paragraph{Cross-Domain Generalization.}
On the MMAR benchmark, which evaluates audio understanding across diverse domains, UAS-Audio achieves the highest overall score of 60.1\%. 
Notably, UAS-Audio demonstrates consistent improvements in both speech-centric tasks (66.0\%, +6.1\% over Qwen2.5-Omni) and music understanding (45.2\%, +4.4\% over Qwen2.5-Omni), suggesting that our unified schema effectively captures domain-specific acoustic attributes. The improvement in speech tasks is particularly significant, as it indicates that enriching supervision with paralinguistic information enhances the model's ability to comprehend speaker characteristics and spoken content simultaneously.

\paragraph{Balanced Audio Understanding.}
On MMAU, while Step-Audio2 achieves the highest overall score (72.7\%), UAS-Audio obtains competitive performance (69.4\%) with notably balanced scores across all three categories (Speech: 67.0\%, Sound: 70.0\%, Music: 71.3\%). This balance performance contrasts with baseline models that exhibit larger variance across domains, demonstrating that UAS provides a more uniform understanding of diverse audio types rather than excelling in specific categories at the expense of others.

\paragraph{Universality Across Architectures.}We further validate the robustness of our approach by applying it to discrete audio representations (Top block of Table~\ref{tab:final_unified_results}). Compared to the baseline GLM-4-Voice, our UAS-Audio-D achieves a transformative improvement, nearly doubling the overall average score (24.4\% $\rightarrow$ 44.2\%). This result confirms that the structured supervision of UAS is effective not only for high-fidelity continuous encodings but also for discrete token-based models.

\subsection{Ablation}
\label{subsection-ablation}
\begin{figure}[t] 
    \centering
    \includegraphics[width=\columnwidth]{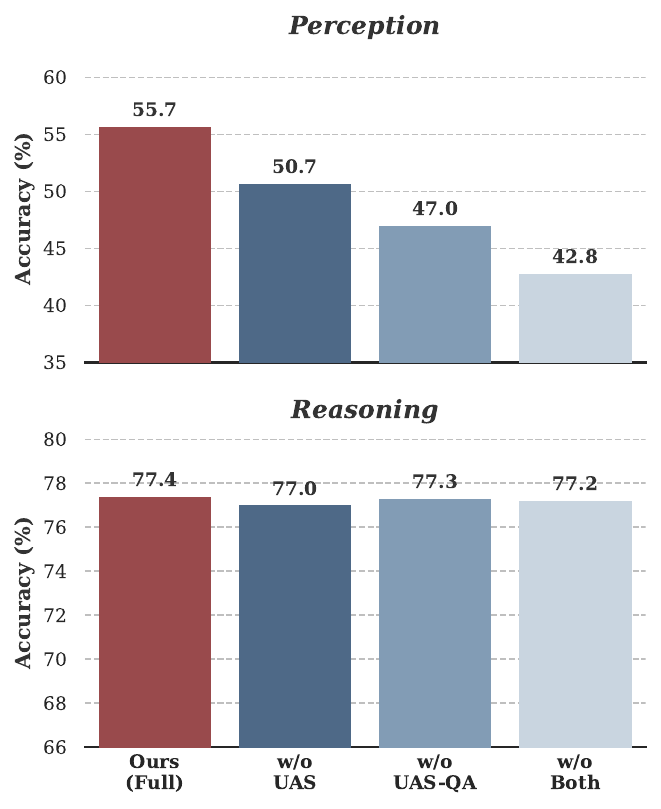}
    \caption{\textbf{Impact of individual components}. We observe a consistent performance drop when removing the UAS module or the UAS-QA strategy from the full model. }
    \label{fig:ablation}
\end{figure}

Figure~\ref{fig:ablation} shows ablation results on MMSU. Both UAS annotation and UAS-QA contribute substantially to perception: removing UAS drops accuracy by 6.3\%, removing UAS-QA by 9.6\%, and removing both by 15.0\%. The larger impact of UAS-QA suggests that explicit question-answering training is more critical for translating acoustic knowledge into task performance.

Crucially, reasoning accuracy remains stable across all configurations, confirming that our perception enhancements do not compromise reasoning capabilities. This validates our hypothesis that perception and reasoning operate as independent factors in audio understanding, and that current Audio LLMs suffer primarily from a perception bottleneck rather than reasoning limitations.

Due to space limitations, further analyses in the appendices ablate the impact of GRPO (Appendix~\ref{app:impact_grpo}) and demonstrate that our structured schema significantly outperforms unstructured captions (Appendix~\ref{app:impact_format}).

\subsection{Speech Generation Ability}
To verify that perception-focused training does not compromise generation capabilities, we evaluate TTS performance on Seed-TTS benchmarks (Table~\ref{tab:tts_results}). UAS-Audio achieves the best average WER score of 1.6, outperforming Qwen2.5-Omni (1.9) and Step-Audio2-mini (2.7). Notably, UAS-Audio matches or exceeds baselines on both Chinese and English synthesis, demonstrating that our UAS training not only preserves but enhances speech generation quality. We attribute this to the enriched acoustic representations learned through structured paralinguistic supervision, which naturally transfer to benefit generation tasks. This confirms that UAS-Audio achieves unified audio intelligence without understanding-generation trade-offs.

\begin{table}[t]
\centering

\setlength{\tabcolsep}{5pt}
\renewcommand{\arraystretch}{1}

\resizebox{0.95\columnwidth}{!}{%
\begin{tabular}{l ccc}
\toprule
\textbf{Model} & Seed-Zh & Seed-EN & \textbf{Avg} \\
\midrule
Baichuan-Audio-Instruct & 2.9 & 4.7 & 3.8 \\
Qwen2.5-Omni        & 1.4 & 2.3 & 1.9 \\
Step-Audio2-mini    & 2.1 & 3.2 & 2.7 \\
\midrule
\textbf{UAS-Audio} & \textbf{1.4} & \textbf{1.7} & \textbf{1.6} \\

\bottomrule
\end{tabular}%
}
\caption{Evaluation results on TTS benchmarks. \textbf{Avg} reports the average score of Seed-Zh and Seed-EN. \textbf{Bold} denotes the best result.}
\label{tab:tts_results}
\end{table}

\subsection{Flexibility and Robustness of Structured Generation.}

While UAS-Audio supports standard, low-latency ASR generation (Targeted Mode), it uniquely possesses the capability to output holistic UAS JSONs---providing rich acoustic context alongside the transcript. A natural question is whether embedding transcription within such a complex structure imposes a "tax" on recognition accuracy. To investigate this, we treat the holistic UAS generation as a stress test and compare its transcription field against standard ASR prompting.

As shown in Table~\ref{tab:asr_robustness}, the transcription quality remains virtually identical across both settings. On the LibriSpeech test-clean set\cite{panayotov2015librispeech}, the WER difference is merely 0.1 (2.2 vs. 2.3), and similarly on AISHELL~\cite{bu2017aishell}, the difference is also 0.1 (2.3 vs. 2.4). This negligible deviation demonstrates that our model maintains high recognition accuracy even when simultaneously predicting paralinguistic attributes, speaker characteristics, and non-linguistic events within a structured JSON format.

%% file: related_work.tex
\label{sec:related}

The evolution of LLMs has driven the transition of spoken dialogue models from traditional cascaded pipelines to end-to-end Audio-Language Models (AudioLLMs)~\citep{zhang2019using,zhang2020graph,zhang2023speechgptempoweringlargelanguage,gong2024listenthinkunderstand,tang2023salmonn,hu2024wavllm,fang2024llama,defossez2024moshi,li2025baichuan,wang2024freeze,bai2024audiosetcapsenrichedaudiocaptiondataset,goel2025audioflamingo3advancing,wijngaard2025audsemthinkerenhancingaudiolanguagemodels}. 
These models generally adopt one of two audio representation strategies: continuous representations ~\citep{huang2025step} or discrete representations~\citep{van2017neural,zeng2024glm,song2026stabletoken}. 
Despite significant progress in speech recognition and dialogue, most existing AudioLLMs primarily focus on \textit{what} is spoken while underutilizing paralinguistic information about \textit{how} it is spoken. 

Recent works have explored richer audio representations. MiDashengLM~\citep{dinkel2025midashenglm} proposes using general audio captions to fuse speech transcripts with acoustic descriptions into a unified textual output. However, this approach relies on \textit{unstructured} natural language, which inherently entangles paralinguistic nuances with linguistic content rather than isolating them. 
In contrast to such implicit supervision, our work advocates for an explicitly structured schema to ensure precise disentanglement of perception and reasoning. SenseVoice~\citep{an2024funaudiollmvoiceunderstandinggeneration} implements rich transcription through interleaved tags, inserting special tokens (e.g., \texttt{<laughter>}, \texttt{<happy>}) directly into the linear ASR text stream. While useful for sparse events, this "flat" tagging structure generalizes poorly to dense, continuous attributes such as prosody, timbre, and emotion. Additionally, it treats these representations as specialized outputs rather than a general-purpose supervision methodology for AudioLLM training.


\begin{table}[t]
\centering
\resizebox{\linewidth}{!}{
\begin{tabular}{l|c|cc}
\toprule
\textbf{Prompt Strategy} & \textbf{Output} & \textbf{LS-Clean} & \textbf{AISHELL} \\
\midrule
Targeted ASR & Raw Text & 2.2 & 2.3 \\
\textbf{Holistic UAS} & \textbf{JSON} & \textbf{2.3} & \textbf{2.4} \\
\midrule
\multicolumn{2}{l|}{\textit{Performance Deviation}} & \textcolor{blue}{\textbf{$\approx$ 0.1}} & \textcolor{blue}{\textbf{$\approx$ 0.1}} \\
\bottomrule
\end{tabular}
}

\caption{\textbf{Robustness of ASR Capability under Multi-Task Generation.} Comparing standard ASR generation versus extracting transcription from the unified JSON output (Holistic). The model maintains high recognition accuracy even when handling complex structured predictions.}
\label{tab:asr_robustness}
\end{table}

%% file: conclusion.tex
We propose the Unified Audio Schema (UAS) to address the deficiency of fine-grained perception in current AudioLLMs. By restructuring supervision across linguistic, paralinguistic, and environmental dimensions, UAS-Audio achieves a 10.9\% improvement in perception accuracy on the MMSU benchmark. Importantly, this gain is achieved without compromising reasoning or transcription capabilities. Our findings suggest that structural richness in supervision is a critical factor for evolving AudioLLMs beyond simple scaling.

%% file: appendix.tex
\cleardoublepage

\section{Human Evaluation of UAS Data Quality}
\label{app:human_eval}

To rigorously assess the reliability of the automated UAS generation pipeline, we conducted a manual quality audit on a randomly sampled subset of the generated dataset. 

\paragraph{Data Sampling and Participants}
We randomly sampled $N=400$ audio segments, ensuring a representative distribution across diverse audio types, including speech, music, and environmental sounds. The evaluation involved three volunteer human annotators with expertise in audio analysis. All participants provided informed consent prior to the task. They were informed that the task involved listening to standard audio samples and that they could withdraw at any time. No monetary compensation was provided.

\begin{figure}[h]
    \centering
    \includegraphics[width=\linewidth]{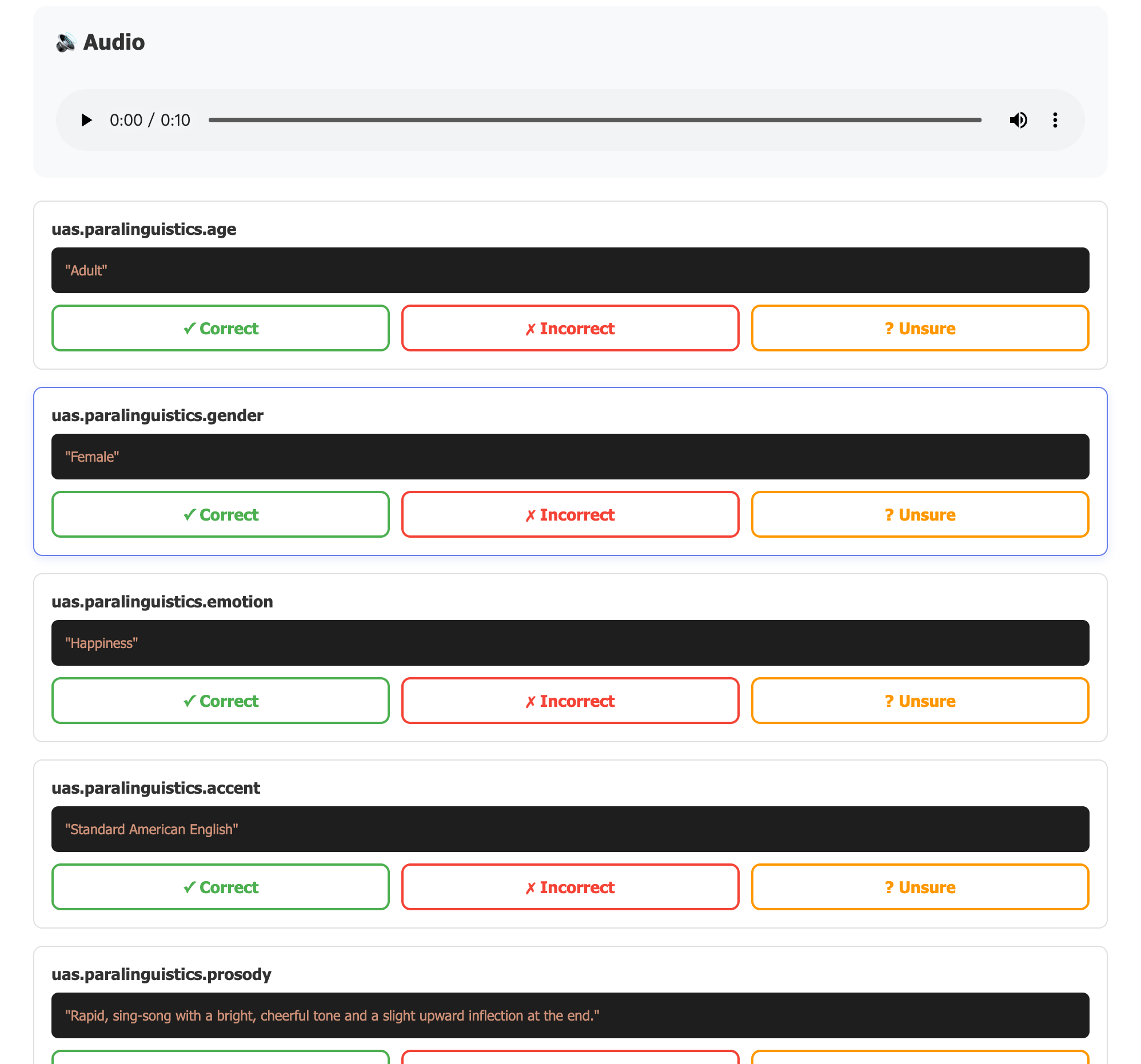} 
    \caption{The web-based human evaluation interface. Annotators listen to the audio and verify if the predicted JSON values (black boxes) match the acoustic reality by selecting "Correct", "Incorrect", or "Unsure".}
    \label{fig:annotation_ui}
\end{figure}

\paragraph{Annotation Interface and Instructions}
To facilitate the evaluation process, we developed a web-based annotation interface, as shown in Figure~\ref{fig:annotation_ui}. The interface presents the audio player at the top, followed by a series of predicted attributes (e.g., Age, Gender, Emotion) derived from the synthesized UAS JSON.

The instructions provided to the participants are as follows:

\begin{quote}
\textit{Please listen to the audio clip provided in the player. Below the player, you will see a list of attributes extracted from the system. Please determine if the label accurately describes the audio content by selecting one of the options: Correct, Incorrect, or Unsure.}
\end{quote}

\paragraph{Evaluation Metric}
For each sample, three annotators independently verified the alignment between the audio signal and the synthesized UAS JSON fields. An attribute was marked as "correct" only if the annotator consensus (majority vote) confirmed it accurately reflected the acoustic reality. We report the mean accuracy across the nine specific fields within the Paralinguistics and Non-linguistic Events domains.

\paragraph{Quantitative Results}
As shown in Table~\ref{tab:human_eval_results}, the UAS pipeline achieves high precision across most fields, with most paralinguistic and environmental attributes exceeding 95\% accuracy. 


\begin{table}[h]
\centering
\resizebox{\linewidth}{!}{
\begin{tabular}{@{}llcc@{}}
\toprule
\textbf{UAS Domain} & \textbf{Field} & \textbf{Accuracy (\%)} & \textbf{95\% CI} \\ \midrule
\multirow{6}{*}{Paralinguistics} & Age & 98.50 & $[96.77, \, 99.31]$ \\
 & Gender & 95.25 & $[92.70, \, 96.94]$ \\
 & Emotion & 89.00 & $[85.55, \, 91.70]$ \\
 & Accent & 96.00 & $[93.60, \, 97.52]$ \\
 & Prosody & 96.50 & $[94.21, \, 97.90]$ \\
 & Timbre & 95.50 & $[93.00, \, 97.13]$ \\ \midrule
\multirow{3}{*}{Non-linguistic Events} & Description & 96.50 & $[94.21, \, 97.90]$ \\
 & Discrete Events & 91.75 & $[88.64, \, 94.07]$ \\
 & Continuous Events & 96.25 & $[92.91, \, 97.71]$ \\ \bottomrule
\end{tabular}
}
\caption{Human evaluation accuracy across UAS fields ($N=400$). 95\% Confidence Intervals (CI) are calculated using the Wilson score interval method.}
\label{tab:human_eval_results}
\end{table}

\paragraph{Analysis} The evaluation reveals that stable biological and environmental traits (e.g., \textit{Age}, \textit{Gender}, and \textit{Continuous Events}) exhibit a high degree of reliability, with lower bounds of the 95\% confidence intervals consistently remaining above 92\%. This suggests that the pipeline effectively captures these relatively objective acoustic properties. The relative performance decrease in \textit{Emotion} (89.0\%, 95\% CI: [85.55\%, 91.70\%]) and \textit{Discrete Events} (91.75\%, 95\% CI: [88.64\%, 94.07\%]) reflects the inherent subjectivity of emotional state perception and the temporal sparsity of short-duration sound events. However, even accounting for statistical uncertainty, the accuracy for these challenging fields remains robustly above 84\%, demonstrating that the UAS pipeline provides a high-fidelity representation across all acoustic dimensions.

\section{Discrete Architecture Training}
\label{app:discrete_training}

For the discrete architecture, we initialize the model with Qwen2.5-3B and adopt the same audio tokenizer as GLM-4-Voice~\citep{zeng2024glm}.

Compared to the continuous architecture, the training pipeline is simplified in two aspects.

First, we omit the adapter alignment stage, since discrete audio tokens are directly embedded into the LLM vocabulary and therefore do not require an additional projection or alignment module.

Second, we do not employ GRPO-style reinforcement learning. This choice follows the training design of GLM-4-Voice, where discrete audio representations are learned purely through supervised objectives without reinforcement learning.

The model is trained on UAS annotation tasks to establish structured acoustic understanding, followed by fine-tuning on UAS-QA tasks to enhance its ability to answer questions about specific acoustic attributes. The learning rate is scheduled with cosine decay, starting from $1 \times 10^{-4}$ and $5 \times 10^{-5}$ respectively.

\section{Training Datasets for UAS-Audio}
\label{app:datasets}
We train UAS-Audio on hundreds of thousands of hours of audio data, including approximately 90\% open-source data and 10\% in-house data. All open-source datasets used in this work are listed in Table~\ref{tab:uas_audio_datasets}.

\begin{table}[ht]
\centering
\resizebox{\linewidth}{!}{
\begin{tabular}{lrll}
\toprule
\textbf{Dataset} & \textbf{Duration (\#hours)} \\
\midrule
LibriSpeech~\citep{panayotov2015librispeech} & 960 \\
Multilingual LibriSpeech~\citep{pratap2020mls} & 27,322 \\
GigaSpeech~\citep{chen2021gigaspeech} & 10,000 \\
Yodas~\citep{li2023yodas} & 29,155 \\
Hi-Fi TTS~\citep{bakhturina2021hi} & 292 \\ 
VCTK~\citep{Veaux2017CSTRVC} & 44 \\
LibriTTS~\citep{zen2019libritts} & 586 \\
AISHELL-1~\citep{bu2017aishell} & 150  \\
WenetSpeech~\citep{zhang2022wenetspeech} & 10,005  \\
Common Voice~\citep{ardila2019common} & 2,133 \\
Emilia~\citep{he2024emilia} & 96,750 \\
AudioSet~\citep{audioset} & 4,922 \\
\bottomrule
\end{tabular}
}
\caption{Summary of datasets used for training UAS-Audio}
\label{tab:uas_audio_datasets}
\end{table}

\section{Inference Efficiency and Flexible Generation Strategy}
\label{app:efficiency}

A potential concern regarding the adoption of the Unified Audio Schema (UAS) is the inference latency and token overhead associated with generating verbose JSON structures. While the full UAS format (containing keys for transcription, paralinguistics, and events) involves a larger number of output tokens compared to simple tagging, it is crucial to distinguish between UAS as a \textit{supervision objective} during training and UAS as an \textit{inference format} during deployment.

\paragraph{Decoupling Supervision from Generation.} 
The primary goal of training with UAS is to force the model to internally disentangle and encode fine-grained acoustic information that is typically suppressed by ASR-centric objectives. Once the model is aligned via this structured supervision (specifically after Stage 3: Full Instruction Tuning), the rich acoustic representations are embedded within the model's parameters. Consequently, during inference, the generation format is entirely flexible and controlled by the user prompt.

\paragraph{Targeted Perception via Prompting.} 
As demonstrated by the inclusion of the UAS-QA dataset in our training pipeline, UAS-Audio is capable of following diverse instructions. For latency-sensitive applications, users need not generate the full holistic JSON. Instead, they can prompt the model to extract specific attributes directly. 

For instance, to retrieve emotion and gender, a user can prompt: \textit{"Identify the speaker's emotion and gender in a concise format."} The model, having learned the underlying concepts through the UAS schema, can output a short response (e.g., \textit{"Neutral, Male"}) without the overhead of the full JSON syntax. This "Targeted Mode" drastically reduces token consumption to be comparable with standard classification heads, while still benefiting from the superior perception accuracy gained from the UAS training paradigm. 

In summary, the structured JSON serves as a high-density information scaffold during learning, but the model retains the flexibility to operate in a low-latency, token-efficient manner during inference.

\section{Impact of GRPO}
\label{app:impact_grpo}
To explicitly quantify the contribution of the reinforcement learning stage, we evaluate a variant of UAS-Audio trained without Stage 4 (GRPO). As presented in Table \ref{tab:grpo_ablation}, removing the GRPO stage results in a marginal performance drop of 0.9\% in perception (55.7\% $\rightarrow$ 54.8\%) and 1.4\% in reasoning (77.4\% $\rightarrow$ 76.0\%).While these results confirm that GRPO serves as an effective  strategy for improving model performance, the ablation highlights a critical finding: even without GRPO, UAS-Audio achieves a perception accuracy of 54.8\%, which still surpasses the strongest baseline (Kimi-Audio, 44.8\%) by a significant margin of 10.0\%. This empirical evidence demonstrates that the substantial performance leap (+11\%) reported in our main results is primarily driven by the structured supervision of the Unified Audio Schema (UAS), rather than the optimization techniques in the final training stage.

\begin{table}[h]
\centering
\resizebox{0.98\linewidth}{!}{
\begin{tabular}{l|ccc}
\toprule
\textbf{Model Settings} & \textbf{Perception} & \textbf{Reasoning} & \textbf{Average} \\ 
\midrule
\textbf{UAS-Audio (Full)} & \textbf{55.7} & \textbf{77.4} & \textbf{66.2} \\
\quad $-$ w/o GRPO (Stage 4) & 54.8 & 76.0 & 65.2 \\
\midrule
\textit{\textcolor{gray}{Best Baseline (Kimi-Audio)}} & \textit{\textcolor{gray}{44.8}} & \textit{\textcolor{gray}{75.7}} & \textit{\textcolor{gray}{62.2}} \\
\bottomrule
\end{tabular}
}
\caption{Ablation study on the impact of GRPO training (Stage 4) on the MMSU benchmark. The results show that while GRPO provides further optimization, the majority of the performance gain stems from the UAS supervision.}
\label{tab:grpo_ablation}
\end{table}

\section{Impact of Structured Format}
\label{app:impact_format}

To validate the effectiveness of our schema design, we conducted a controlled comparison between \textbf{UAS Supervision} (structured JSON) and a \textbf{Caption Supervision} baseline (unstructured natural language). 
Notably, this Caption Supervision setting conceptually emulates the \textit{general audio captioning} paradigm adopted by recent works such as MiDashengLM~\citep{dinkel2025midashenglm}.
We exclude the final reinforcement learning stage (Stage 4) in both settings to strictly isolate the impact of the supervision format while controlling for model architecture and training data.

\begin{table}[h]
\centering
\resizebox{\linewidth}{!}{
\begin{tabular}{l|ccc}
\toprule
\textbf{Target Format} & \textbf{Perception} & \textbf{Reasoning} & \textbf{Average} \\
\midrule
Unstructured Caption & 48.4 & 75.5 & 61.5 \\
\textbf{Structured UAS} & \textbf{54.8} & \textbf{76.0} & \textbf{65.2} \\
\bottomrule
\end{tabular}
}
\caption{Impact of supervision format on MMSU. The "Unstructured Caption" setting serves as a proxy for caption-based approaches (e.g., MiDashengLM). Both settings use the exact \textbf{same synthetic data source}. All models are trained without GRPO.}
\label{tab:ablation_schema}
\end{table}

As shown in Table~\ref{tab:ablation_schema}, structured UAS yields a significant 6.4\% perception gain over the unstructured caption baseline. confirming that the schema format itself lowers learning difficulty given identical data. 
Specifically, UAS enforces \textbf{explicit disentanglement} by assigning orthogonal slots to prevent semantic interference, offers \textbf{syntactic invariance} by providing a low-entropy target devoid of linguistic variability, and ensures \textbf{forced completeness} by mandating dense predictions for all acoustic fields, thereby capturing subtle details often omitted in fluent captions.

\section{Training Hyperparameters}
\label{app:hyperparameters}

Table~\ref{tab:hyperparameters} summarizes the hyperparameter configurations for each training stage.

\begin{table*}[h]
\centering
\begin{tabular}{lcccc}
\toprule
\textbf{Hyperparameter} & \textbf{Stage 1} & \textbf{Stage 2} & \textbf{Stage 3} & \textbf{Stage 4} \\
\midrule
Peak Learning Rate & $5\text{e-}4$ & $2\text{e-}4$ & $1\text{e-}4$ & $5\text{e-}6$ \\
Warmup Iterations & 500 & 1,000 & 1,000 & 200 \\
LR Schedule & \multicolumn{4}{c}{Cosine with Linear Warmup} \\
Optimizer & \multicolumn{4}{c}{AdamW ($\beta_1=0.9$, $\beta_2=0.95$)} \\
Weight Decay & \multicolumn{4}{c}{0.1} \\
Gradient Clipping & \multicolumn{4}{c}{1.0} \\
\midrule
Trainable Parameters & Projector & Projector & All (excl. Encoder) & All (excl. Encoder) \\
\bottomrule
\end{tabular}
\caption{Hyperparameter configurations across training stages.}
\label{tab:hyperparameters}
\end{table*}

\section{Prompts Used in Experiments}
\label{app:prompts}

In this section, we provide the detailed prompts used in our pipeline for reproducibility.

\subsection{Audio Caption to UAS Format Conversion}

We utilized the Qwen3-30B-A3B-Instruct model to convert raw audio captions into the Unified Audio Description (UAS) format. The prompt used for this transformation is shown in Figure~\ref{fig:prompt1}.

\begin{figure*}[h!]
    \begin{promptbox}{Prompt for Caption-to-UAS Conversion}
Given a detailed description of an audio sample, output a JSON object containing the following audio features:

- **transcription**: If human speech is present, provide an accurate transcription of the spoken content in the original language. If there is no human voice, set this field to null.
- **paralinguistics**: If human voice is present, provide the following fields:  
  - `age`: One of `Child`, `Adult`, or `Elderly`.
  - `gender`: Specify as `Male` or `Female`.
  - `emotion`: This field MUST use ONE of the following seven specific categories: `Anger`, `Disgust`, `Sadness`, `Happiness`, `Neutral`, `Surprise`, `Fear`. Only these values are allowed.
  - `accent`: Describe the accent or variety of language used (e.g., `Standard Mandarin Chinese`, `American English`, etc.).
  - `prosody`: Summarize prosodic features, which refer to the patterns of rhythm, pitch, pace, emphasis, and intonation in speech (i.e., how something is said).
  - `timbre`: Briefly describe the timbre of the voice. **Timbre** refers to the unique tonal quality or color of a sound that distinguishes one voice or instrument from another, independent of pitch and loudness. For example, descriptors may include "nasal," "breathy," "warm," "bright," "harsh," or "gentle."

  **Note:** Timbre is *not* the same as prosody; prosody relates to temporal and pitch-based features, while timbre describes the characteristic sound qualities.

  If there is no human voice, set all fields in the `paralinguistics` object to null.
- **nonLinguisticEvents**:
  - `description`: A summary sentence describing general non-speech audio characteristics or context.
  - `discreteEvents`: A list of discrete (one-shot or instantaneous) non-linguistic events (such as a car horn, a door slam). Each item must contain a unique `label` and a brief `characteristic` describing its intensity, duration, or other relevant attribute. (e.g., `label`: `"Car horn"`, `characteristic`: `"Short, loud"`). Event labels must not repeat.
  - `continuousEvents`: A list of continuous or background non-linguistic events (such as engine noise, wind, music), again with a unique `label` and a brief `characteristic` descriptor.

Always follow these rules:
- If the audio contains **no human voice**, set `transcription` and all fields inside `paralinguistics` to null.
- For `emotion`, ONLY USE ONE OF THESE: `Anger`, `Disgust`, `Sadness`, `Happiness`, `Neutral`, `Surprise`, `Fear`.
- Ensure that all event `labels` are unique and clearly indicate what type of sound or event they refer to.

Respond ONLY with a JSON object as output (do not include any preamble, explanation, or extra formatting), with all required fields. Use the formats and categories exactly as described above.
\end{promptbox}
\caption{Prompt for using the Qwen3-30B-A3B-Instruct model to perform Caption-to-UAS Conversion}
\label{fig:prompt1}
\end{figure*}

\subsection{QA Pair Generation from UAS}

To generate Question-Answer (QA) pairs based on the audio UAS format descriptions, we employed the Qwen3-235B-A22B-Instruct model. The specific prompt guiding this generation process is detailed in Figure~\ref{fig:prompt2}.

\begin{figure*}[h!]
\begin{promptbox}{Prompt for QA Generation}
**Instructions:**
You are given a structured audio description in UAS (Unified Audio Schema) JSON format. Please generate a relevant question in the form of a **Multiple Choice** question, along with the corresponding answer, based on the specific fields provided in the JSON (such as transcription, paralinguistics, or non-linguistic events).

**Requirements:**
- Provide 3-4 answer options. Each option must include both the letter and the content (e.g., "A. male", "B. female").
- The question can pertain to specific attributes found in the UAS structure, such as:
  - The speaker's gender, age, emotion, accent, prosody, and timbre (from `paralinguistics`).
  - Specific sounds or events (from `discreteEvents` or `continuousEvents`).
  - The content of speech (from `transcription`).
- The question text must not directly reveal or hint at the answer; answering must require information from the audio, and not be possible by simply reading the question.
- Do not include phrases like "according to the JSON" or "in the paralinguistics field".
- The correct answer must be option ${correct_option}.

**Input Format:**
A JSON object containing `transcription`, `paralinguistics`, and `nonLinguisticEvents`.

**Output Format:**
Present your output in the following JSON format:
```json
[
    {"role": "user", "content": [{"type": "text", "text": "question_text"}]},
    {"role": "assistant", "content": "answer_text"}
]
```

**Now, generate a question and its answer for the following UAS input using the above guidelines:**

${uas}

\end{promptbox}
\caption{Prompt for using the Qwen3-235B-A22B-Instruct model to perform QA Generation from UAS input}
\label{fig:prompt2}
\end{figure*}

\section{LLM Usage Statement}
In accordance with the conference policies on Large Language Model (LLM) usage, we hereby disclose the following: After completing the initial draft of this paper, we utilized LLMs to enhance grammar and polish the writing of this manuscript. No new research ideas, experimental designs, or scientific content were generated by LLMs. 

This statement is provided to ensure transparency and compliance with the conference’s policies on
LLM usage.